\documentclass[letterpaper]{article} 
\usepackage{aaai25}  
\usepackage{times}  
\usepackage{helvet}  
\usepackage{courier}  
\usepackage[hyphens]{url}  
\usepackage{graphicx} 
\usepackage{natbib}  
\usepackage{caption} 
\usepackage{algorithm}
\usepackage{algorithmic}

\usepackage{newfloat}
\usepackage{listings}
\DeclareCaptionStyle{ruled}{labelfont=normalfont,labelsep=colon,strut=off} 
\floatstyle{ruled}
\newfloat{listing}{tb}{lst}{}
\floatname{listing}{Listing}
\title{Model Misalignment and Language Change:\ Traces of AI-Associated Language in Unscripted Spoken English}
\author{
    Bryce Anderson\footnotemark[1], Riley Galpin, Tom S.\ Juzek\thanks{\noindent Corresponding authors. Author contributions:\ Conceptualisation and methodology (BA, RG, TSJ), data collection and analysis (BA, TSJ), coding and visualisation (BA, TSJ), write up (BA, RG, TSJ).}
}
\affiliations{
    Florida State University\\

}

\usepackage{bibentry}

\begin{document}

\maketitle

\begin{abstract}
In recent years, written language, particularly in science and education, has undergone remarkable shifts in word usage. These changes are widely attributed to the growing influence of Large Language Models (LLMs), which frequently rely on a distinct lexical style. Divergences between model output and target audience norms can be viewed as a form of misalignment. While these shifts are often linked to using Artificial Intelligence (AI) directly as a tool to generate text, it remains unclear whether the changes reflect broader changes in the human language system itself. To explore this question, we constructed a dataset of 22.1 million words from unscripted spoken language drawn from conversational science and technology podcasts. We analyzed lexical trends before and after ChatGPT's release in 2022, focusing on commonly LLM-associated words. Our results show a moderate yet significant increase in the usage of these words post-2022, suggesting a convergence between human word choices and LLM-associated patterns. In contrast, baseline synonym words exhibit no significant directional shift. Given the short time frame and the number of words affected, this may indicate the onset of a remarkable shift in language use. Whether this represents natural language change or a novel shift driven by AI exposure remains an open question. Similarly, although the shifts may stem from broader adoption patterns, it may also be that upstream training misalignments ultimately contribute to changes in human language use. These findings parallel ethical concerns that misaligned models may shape social and moral beliefs. 
\end{abstract}

\begin{links}
     \link{GitHub repository for code and data}{https://github.com/fsu-nlp/ai-traces-ssl}
\end{links}

\section{Introduction}

\begin{figure}[ht]
    \centering
    \includegraphics[width=1\columnwidth]{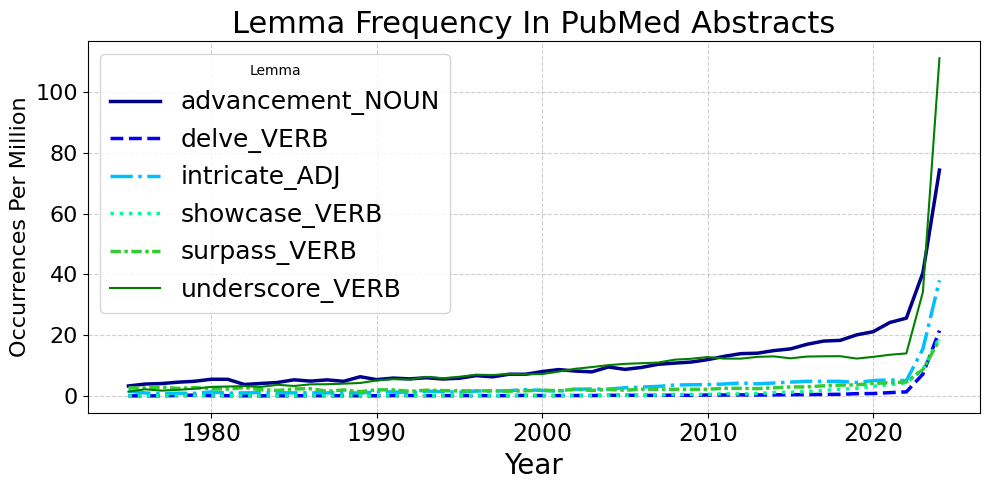}
    \caption{Whether the sharp post-2022 rise in AI-associated words reflects mere tool usage or a direct influence on the human language system remains an open question with broader societal relevance.}
    \label{fig:hook}
\end{figure}

Languages change over time, and the processes underlying language change are complex, involving a combination of social \cite{Amato_2018, zhang2014review}, cultural \cite{labov2011principles}, and psycho-cognitive factors \cite{ClossTraugott1985,li2023cognitive}. Over the past centuries, technology has also played a significant role in driving language change. The introduction of the printing press, for example, expanded access to written literature, promoted the standardization of written language, and transformed global communication \cite{Elizabeth1980languageevolution}. Similarly, the invention of the telephone and later the cell phone influenced the way humans speak and interact with one another \cite{crystal2008txtng, Sandra2011influenceofcomputer, alsharqi2020influence,mamatkulova2025influence}, similarly the internet in general \cite{Helen2024effectofinternet}, as well as social media \cite{vanisree2024effects}. Language is the medium that humans use to communicate ideas \cite{clark1996using,fedorenko2024language}. As such, understanding  the possibilities and constraints of language, as well as understanding language change and the processes behind it matters. 

Selected rapid shifts in word usage do occur and are typically traceable to real-world events (``Omicron,'' ``Türkiye''). This contrasts with the recent, large-scale shifts observed in certain domains, particularly in education and science, which appear to not be triggered by external world events. Words such as ``delve,'' ``intricate,'' and ``underscore'' have seen sudden spikes in usage, especially within academic writing (\citealp{gray2024chatgpt,kobak2024delving,liang2024mapping,liu2024towards,matsui2024delving,comas2025ai,bao2025examining}; see also Figure~\ref{fig:hook}, which shows the frequency of selected lemmata, i.e.,\ base forms of words, in PubMed abstracts). Scripted spoken language, such as academic presentations, has been analyzed and similar trends were observed \cite{geng2024impact,yakura2024empirical}. 
These changes are often attributed to the widespread introduction of Large Language Models (LLMs) with chat functionality, especially ChatGPT \cite{OpenAI2024}, which overuses words such as ``delve'' and ``intricate'' \cite{gray2024chatgpt,Koppenburg2024,kobak2024delving,juzek2025does}, and exhibits a distinct syntactic style \cite{zamaraeva2025comparing}. Further, these linguistic preferences are frequently discussed in the context of model alignment:\ is there a gap between model behavior and user expectations \cite{hendrycks2020aligning,gabriel2020artificial,norhashim2024measuring}, and if so, where does it originate? One possibility is that such misalignments arise during the training phase, particularly through the process of Learning from Human Feedback (in the literature also referred to as Human Preference Learning and Reinforcement Learning from Human Feedback; the latter term, however, is not ideal, as not all preference procedures involve reinforcement learning). \citet{zhang2024lists} argue that the overuse of stylistic elements like boldfacing originates from Learning from Human Feedback, and \citet{juzek2025does} suggest that lexical overuse may similarly arise during this phase. However, if such biases reflect human preferences (just not those of the intended user base) rather than contradict them, the concept of ``misalignment'' becomes ambiguous (cf.\ \citealp{blodgett2020language} on the ambiguity of such definitions). 

Crucially, these issues mirror broader debates around model alignment in the domains of ethics and fairness \cite{russell2015research,gabriel2020artificial,perez2023discovering, hendrycks2020aligning, norhashim2024measuring}, where biases related to race, ethnicity, nationality \cite{narayanan2023unmasking,omiye2023large, akintande2023algorithmic, ali2023evaluating, naik2023social}, intersectionality \cite{omrani2023evaluating}, and gender \cite{bolukbasi2016man,ghosh2023chatgpt,kotek2023gender,wilson2024gender, ali2023evaluating, naik2023social, jaiswal2024uncovering} are often amplified not because models deviate from human input, but because they replicate dominant patterns already present in it \cite{grabowicz2023learning, akintande2023algorithmic}. As such, misaligned models can cause real harm \cite{neff2016talking,bender2021dangers,hagen2025grok}. Thus, insights into LLMs' language use have the potential to inform more general alignment discussions about whose preferences, whether it would be values or language preferences, are being encoded and reinforced.

\subsection{Research question}

When it comes to the observed language change, however, an important piece in the debate is missing:\ it remains an open question whether the observed shifts in language use are simply the result of tool usage (e.g., a particular instance of ``delve'' was generated by ChatGPT, and the human used it), or whether the language choices of LLMs have influenced human speakers and writers to such an extent that the shift reflects an actual change in human language use (i.e., ``delve'' was produced by a human, potentially influenced by prior exposure to LLM output) \cite{clark1996using,fedorenko2024language}. This distinction matters:\ if the influence is direct, it strengthens the case for more robust model alignment, as linguistic preferences encoded by LLMs may begin to shape human communication norms. Thus, the research question we approach is: 

\begin{center}
\textbf{(RQ) Are human lexical choices in unscripted spoken language increasingly converging towards the lexical patterns characteristic of Large Language Models?}
\end{center}

\noindent That is, among the words that LLMs are known to overuse (as documented in the literature; see the Methods subsection for the list of words that we investigated), can we observe an uptick in genuine human language production, excluding tool-generated language? Our hypothesis is that such an uptick exists, at least to some degree. The null hypothesis is that there is no notable increase.

Measuring ``genuine'' human language production presents considerable challenges, especially in written texts. Before the widespread adoption of Large Language Models, the default assumption was that any given text was authored by a human. This presumption no longer holds. For many texts, it is not clear anymore if they were genuinely authored by a human, i.e.\, they present a case of \textbf{human-authorship indeterminacy} (see \citealp{huang2025authorship} and \citealp{sadasivan2023can} for related thoughts). For linguistic analysis aimed at understanding the human language system, this marks a considerable disruption, especially since most linguistic research has focused on the analysis of written language. The field's (over)reliance on written data has been criticized in the past \cite{biber1991variation}, and the rise of AI-generated content makes deriving linguistic insights from such data increasingly difficult. The analysis of spoken language holds value \cite{miller1998spontaneous}; however, even spoken language is not immune, as academic presentations, for instance, are often scripted, and these scripts may be generated by Artificial Intelligence (AI) models. To address this challenge, we focus on unscripted spoken language from conversational podcasts.

\begin{table*}[t]
\centering
\begin{tabular}{|l|c|c|c|c|}
\hline
\textbf{Word} & \textbf{Change (\%)} & \textbf{OPM Pre-2022} & \textbf{OPM Post-2022} & \textbf{p \ensuremath{\leq} 0.05} \\
\hline
surpass\_VERB         & 140.79 & 1.47  & 3.53  & \textbf{True} \\
boast\_VERB           & 140.14 & 0.82  & 1.98  & \textbf{True} \\
meticulous\_ADJ       & 125.52 & 0.46  & 1.03  & False \\
strategically\_ADV    & 87.93  & 1.74  & 3.27  & \textbf{True} \\
garner\_VERB          & 53.76  & 1.01  & 1.55  & False \\
notable\_ADJ          & 53.13  & 2.47  & 3.79  & False \\
intricate\_ADJ        & 47.21  & 2.75  & 4.04  & False \\
delve\_VERB           & 46.82  & 2.93  & 4.30  & False \\
align\_VERB           & 36.59  & 18.77 & 25.64 & \textbf{True} \\
advancement\_NOUN     & 30.27  & 4.03  & 5.25  & False \\
significant\_ADJ      & 17.35  & 39.74 & 46.63 & \textbf{True} \\
showcase\_VERB        & 6.49   & 1.37  & 1.46  & False \\
comprehend\_VERB      & 3.36   & 3.66  & 3.79  & False \\
pivotal\_ADJ          & 0.45   & 2.66  & 2.67  & False \\
potential\_NOUN       & -2.95  & 33.51 & 32.52 & False \\
emphasize\_VERB       & -4.15  & 18.22 & 17.47 & False \\
potential\_ADJ        & -12.15 & 33.79 & 29.68 & False \\
realm\_NOUN           & -17.63 & 22.25 & 18.33 & \textbf{True} \\
crucial\_ADJ          & -24.56 & 19.05 & 14.37 & \textbf{True} \\
groundbreaking\_ADJ   & -40.94 & 3.20  & 1.89  & False \\
\hline
\end{tabular}
\caption{The target words occurring in our dataset, including their percentage change in frequency (occurrences per million; OPM) for pre-2022 vs.\ post-2022, and whether the difference is significant.}
\label{tab:targetwords}
\end{table*}

\subsection{Further related work}

\citet{zhang2024lists} investigate changes in formatting, such as lists and boldface. \citet{geng2024impact} and \citet{yakura2024empirical} examine scripted spoken language. \citet{liang2024mapping} map shifts in word usage in written academic texts, while \citet{pack2023affordances} focus specifically on the domain of education. \citet{galpin2025lexical} offer a deeper structural analysis of word usage patterns and their changes over time. \citet{zhang2024lists} observe that LLMs overuse certain stylistic elements like boldface and making use of lists, and implicate Learning from Human Feedback, a procedure to align LLMs with human preferences \cite{christiano2017deep,ziegler2019fine,ouyang2022training,rafailov2024direct}. \citet{wu-aji-2025-style} observe that humans tend to prefer model outputs with polished style and lower substance over outputs with greater substance but less polished style. \citet{juzek2025does} argue that, with respect to the overuse of certain words, such model behavior also stems from Learning from Human Feedback. We contribute to this body of work by focusing on unscripted spoken language and by directly addressing the question of whether the human language system itself has been affected by the presence and output of AI. 

Beyond linguistic considerations (to give an incomplete selection), further considerations include \citet{norhashim2024measuring}, who propose frameworks for quantitatively assessing human-AI value alignment in Large Language Models. \citet{hendrycks2020aligning} introduce benchmark tasks aimed at assessing alignment with shared human values, including helpfulness, honesty, and harmlessness. \citet{keswani2024pros} explore how active learning methods used for preference elicitation may shape the moral values internalized by models. \citet{leidinger2024llms} contribute to ongoing debates on mitigating stereotyping harms in generative systems, while \citet{welbl2021challenges} and \citet{hartvigsen2022toxigen} discuss challenges with respect to detoxification and adversarial training, respectively. Finally, all these considerations are accompanied by discussions about governance and ethical frameworks, including proposals for algorithmic accountability \citep{mittelstadt2016ethics}, ethically aligned AI \citep{floridi2018ai4people}, and global AI ethics guidelines \citep{jobin2019global}.

\section{Methods}

\textbf{Target words}:\ For the AI-related words under investigation, we used the 34 words identified in \citet{galpin2025lexical}, which were drawn from prior literature and their own analyses. For any of the 34 words, it was established through a formal analysis that there has been a considerable increase in the words' usage pre- and post-ChatGPT's release in 2022. The selection process prioritizes words that either displayed the most pronounced changes or received substantial attention in the literature (ibid.). Out of the 34 target words from the literature, 20 did occur in our dataset. The list of occurring words is provided in Table~\ref{tab:targetwords}. The 14 words not occurring in our dataset are:\ commendable\_ADJ, comprehending\_ADJ, emphasized\_ADJ, garnered\_ADJ, intricacy\_NOUN, invaluable\_ADJ, meticulously\_ADV, noteworthy\_ADJ, showcase\_NOUN, showcasing\_ADJ, surpassing\_ADJ, underscore\_VERB, underscore\_NOUN, underscoring\_ADJ. All of the non-occurring words are low-frequency words (note the infrequent adjectival usage in many of these words), most of which occur well below one instance per million words, for example, when analyzing occurrences in PubMed abstracts. Moreover, all of the 34 words belong to a more formal or academic register, which makes their occurrence in unscripted spoken language less likely. (All scripts, sample data, and results can be found in our GitHub repository.) \\ 

\noindent \textbf{Dataset construction}: Because of the issue of human-authorship indeterminacy, our dataset consists of unscripted spoken language from conversational podcasts. We focused on the most popular podcasts in the United States related to technology and science, based on the assumption that speakers engaging with these topics are likely to be directly or indirectly exposed to LLM-generated language. Accordingly, this subgroup offers a high probability of revealing potential effects. To determine popularity and identify relevant podcast content, we looked at Apple and Spotify algorithm recommendations, as well as queried Google, YouTube, and ChatGPT. All prompts included the condition that results were to return only conversational podcasts within both genres to maximize capturing unscripted language. For each candidate podcast, we listened to a sample episode to verify the presence of spontaneous, unscripted speech.

\begin{table*}[t]
    \centering
    \begin{tabular}{|l|c|c|}
        \hline
        \textbf{Podcast Title} & \textbf{Number of Episodes} & \textbf{Number of Tokens}\\
        & \textbf{Pre-22 \& Post-22} & \textbf{Pre-22 \& Post-22}\\
        \hline
        BBC Curious Cases&  20 \& 20 &  0.13m \&  0.14m\\
        \hline
        Big Picture Science&  15 \& 15&  0.13m \&  0.15m\\
        \hline
        Big Technology Podcast&  40 \& 40&  0.47m \&  0.48m\\
        \hline
        Brain Inspired&  20 \& 20 &  0.30m \&  0.36m\\
        \hline
        Conversations with Tyler&  36 \& 36&  0.44m \&  0.45m\\
        \hline
        CoreC&  14 \& 14 &  0.14m \&  0.16m\\
        \hline
        EconTalk&  64 \& 64&  0.94m \&  0.88m\\
        \hline
        The Ezra Klein Show &  60 \& 60&  0.80m \&  0.77m\\
        \hline
        Lex Fridman&  93 \& 93&  2.25m \&  3.61m\\
        \hline
        Machine Learning Street Talk&  30 \& 30 &  0.69m \&  0.59m\\
        \hline
        Mindscape&  36 \& 36&  0.66m \&  0.59m\\
        \hline
        Ologies with Alie Ward&  67 \& 67&  1.07m \&  1.22m\\
        \hline
        People Behind the Science&  16 \& 16&  0.40m \&  0.15m\\
        \hline
        Radiolab&  61 \& 61&  0.67m \&  0.61m\\
        \hline
        STEM Talk&  29 \& 29&  0.42m \&  0.44m\\
        \hline
        Talk Nerdy with Cara Santa Maria&  31 \& 31&  0.43m \&  0.30m\\
        \hline
        Very Bad Wizards&  30 \& 30 &  0.57m \&  0.66m\\\hline
    \end{tabular}
    \caption{The list of podcasts analyzed for this work, including the number of episodes and token count (in millions) per podcast.}
    \label{tab:podcasts}
\end{table*}

Given that ChatGPT, the most influential chat-based LLM, was released in late 2022, we divided our dataset into two periods:\ pre-2022 (mostly 2019 to 2021) and post-2022 (2023 to 2025). The year 2022 was excluded from the dataset, as it marks the release of ChatGPT and represents a transitional period between the pre- and post-ChatGPT eras. To keep the dataset balanced we collected a 1:1 ratio of pre/post episodes from each podcast. The 1:1 ratio also ensures that any potential regional or varietal differences are controlled for, as each podcast contributes equally to both time periods. Further, we tried to include more episodes from podcasts with longer average runtimes and fewer episodes from podcasts with shorter runtimes, but the counts vary due to a finite amount of podcast episodes that fit within our criteria, and could be downloaded in equal quantities for both pre- and post-2022. When available, we used transcripts provided by the podcast itself. If no transcripts were available, we downloaded the audio and transcribed it using Python \cite{Python3} and OpenAI's Whisper \cite{radford2022whisper}.

We concluded data collection once we had amassed about 11 million words for each time period (22.07 million words in total). In total, we included 1326 episodes, and a complete list of the podcasts included in our study is provided in Table~\ref{tab:podcasts}, including well-known examples such as Lex Fridman and Stem Talk. Due to copyright restrictions, we can only share snippets of the transcripts, which are available on our GitHub. However, our methodology is fully replicable:\ the original audio and transcripts are publicly accessible, and all scripts used in the analysis are also available on our GitHub. The details of our exact technical setup can be found in Appendix~A. \\

\noindent \textbf{Data processing and analysis}: We analyze lemmata, i.e.\ inflected forms are analyzed as a single item (e.g., `delves' and `delved' have the common lemma `delve'). Further, we distinguish part-of-speech (POS) categories (as per \citealt{jurafsky2024slp}); for example, the word ``lead'' may refer to the noun ``lead,'' a physical material, or the verb ``lead,'' which holds an entirely different meaning. Accordingly, we lemmatized and POS-tagged all transcripts using spaCy (\citealp{montani2023spacy}; tagging took about 2 hours). For the analysis, we compare occurrences per million (OPM) across the two time periods:\ pre-2022 and post-2022. Our main analysis assesses whether overall lexical usage has shifted by computing the weighted mean of the log frequency ratios and testing whether this change differs from zero using a z-test. To prevent instability due to low-frequency items, we applied Laplace smoothing by adding 0.5 to all counts before computing the log of the post-/pre-2022 frequency ratio. The weighting accounts for differences in frequency magnitude and statistical reliability across items:\ weights were defined as the inverse of the estimated variance, calculated as 1 / (1/f\_post + 1/f\_pre). To regularize the undue influence of extremely frequent items, we capped all weights at 20. The resulting z-test yields a robust group-level inference \cite{bland2015introduction}.

As a baseline, we also analyze synonyms identified in \citet{galpin2025lexical}. For instance, for ``delve,'' the semantically related verbs ``explore,'' ``investigate,'' ``examine,'' ``research,'' and ``probe'' were identified as synonyms, for which we then examine changes in their use accordingly. A full list of baseline words can be found in Appendix~B. 

We additionally analyze individual lemmata using chi-square contingency table tests, which are appropriate for our data structure \citep{haslwanter2016introduction}, given the independent, unpaired samples and large dataset size. However, because we conduct dozens of tests (especially for the baseline items) and expect a number of true effects in both directions, applying a strict alpha-level correction (e.g., Bonferroni) would be overly conservative and obscure meaningful effects. Thus, the individual results should be interpreted with caution:\ at an uncorrected alpha level of 0.05, some false positives are expected.

\section{Results}

\begin{figure*}[ht]
    \centering
    \includegraphics[width=2\columnwidth]{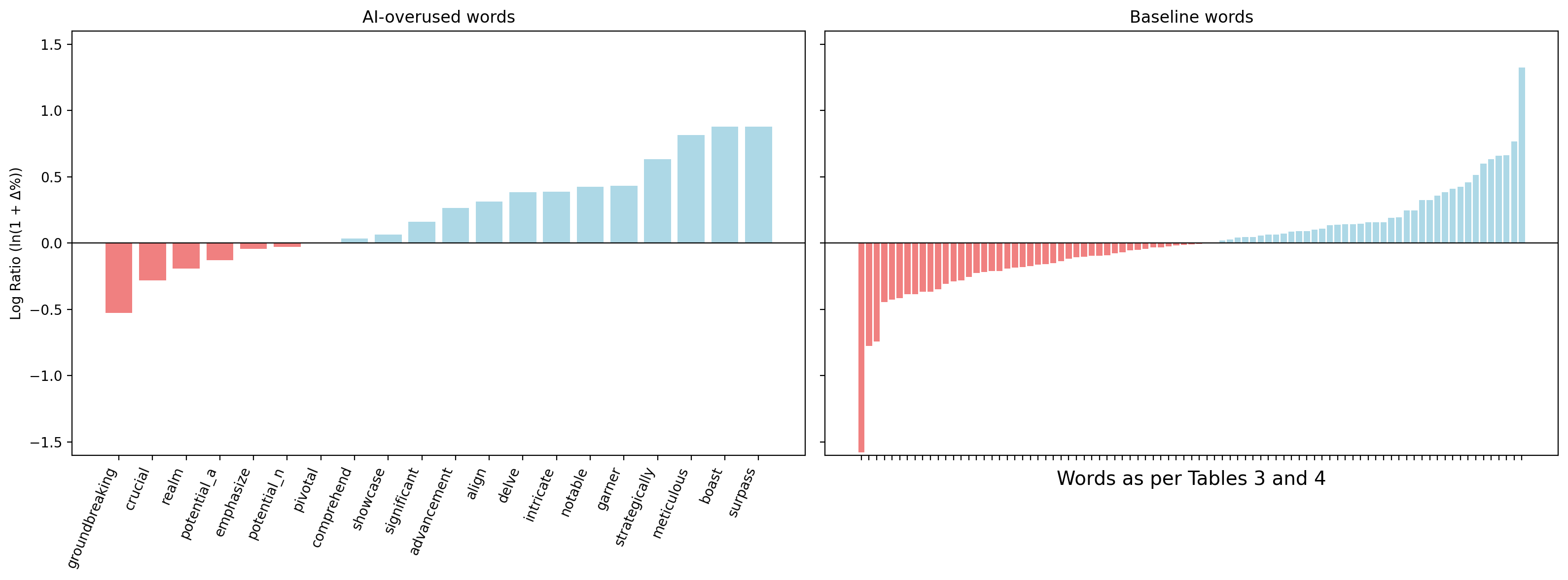}
    \caption{Pre- vs post-2022 logged proportional change for AI-associated words (left) vs baseline words (right).}
    \label{fig:identificationprocedure}
\end{figure*}

The results for the 20 words that we analyzed are presented in Table~\ref{tab:targetwords} and Figure~\ref{fig:identificationprocedure}. Figure~\ref{fig:identificationprocedure} gives the logged proportional change, to account for the fact that proportional increases and decreases are asymmetrical in raw form. For example, a doubling (+100\%) and a halving (-50\%) are not numerically equivalent in raw form, but become symmetric when expressed in logarithmic space.  

For the target words (corresponding to Figure~\ref{fig:identificationprocedure}, left), the weighted mean analysis reveals a moderate overall increase in usage (weighted log-ratio mean \( = 0.210 \)). This change is statistically significant \( (z = 3.725,\ p < 0.001) \).  Looking at the target words individually, we observe that of the 20 target words in our dataset, 6 show a decline in usage between the pre- and post-2022 periods. Of these, 2 changes are statistically significant according to the chi-square contingency table test:\ ``crucial'' and ``realm.'' 14 words show an increase in usage, 5 of which are statistically significant:\ ``significant,'' ``align,'' ``strategically,'' ``boast,'' and ``surpass.'' 

By contrast, the overall analysis of the baseline words (Figure~\ref{fig:identificationprocedure}, right) shows only a very small, non-significant increase (weighted log-ratio mean \( = 0.033,\ z = 1.277,\ p > 0.05 \)). Analyzing the baseline words individually, we observe that out of the 117 synonyms included in our analysis, only 87 occur in our dataset (for the same reason as with our target words:\ several of these items have extremely low frequencies to begin with, even in formal writing). For the majority of these (63 out of 87), changes in frequency are relatively moderate, remaining below +/-30\%. Notably, the overall direction of change is balanced:\ approximately half of the words show a decrease in usage (45 out of 87), while the other half show an increase (42 out of 87). Nonetheless, several words exhibit considerable shifts between the pre- and post-2022 periods, some of which are statistically significant. The full results for the baseline words are provided in Appendix~B.

\section{Discussion}

\subsection{General discussion}

Overall, we observe a significant positive tendency, which stands in contrast to the baseline words, for which changes are roughly balanced between increases and decreases (with only a very small, non-significant upward shift). This allows us to reject the null hypothesis that there is no notable increase. However, the alternative hypothesis must be interpreted with nuance. We observe an overall moderate uptick, with most AI-associated words increasing in usage. This suggests that human word choices have, moderately and selectively, converged with those of Large Language Models. However, given the short time frame and the number of items showing increases, this could signal the onset of a remarkable shift in language use. 

There are several observations to make. First, in contrast to the sharp spikes in written usage of these words (see Figure~\ref{fig:hook}), the effect in spoken language may be less strong than one might have expected. This contrast suggests that the increase in written texts may be largely driven by tool usage, rather than a rapid shift in the human language system (though the latter cannot fully be ruled out, it would be unprecedented in scale and would require substantially stronger evidence). 

Notably, the word ``delve'' did not show a significant increase; its rise was smaller than anticipated. Even more surprising is that ``realm,'' despite being frequently discussed in the literature as a characteristic LLM word (e.g.\ \citealp{liu2024towards}), significantly decreased in usage. Among the words that did show significant increases, ``align'' needs further analysis, as its rise may be linked to a broader shift in discourse, viz.\ the growing conversation around aligning LLMs with human values. 

For the observed upticks, particularly the significant ones such as ``surpass,'' it is tempting to conclude that LLMs have caused these changes. Given that many of the words heavily overused by AI are also showing increased usage in human language, it is plausible to conjecture a link. However, the question of causation remains open. As noted by \citet{matsui2024delving,juzek2025does}, many of the words now considered overused by AI were already trending upward in prior years (see pre-2022 trajectories illustrated in Figure~\ref{fig:hook}). It is possible that these words have simply entered a phase of natural, rapid adoption, akin to the rise of expressions like ``touch base,'' ``dude,'' and ``awesome'' in the mid-2000s. As such, the observed changes could be in line with language change observed in the past (see \citealp{aitchison2005language_change,bochkarev2014lexical_evolution} for descriptions of conventional language change). Under this view, LLMs overuse words that were already increasing in frequency, patterns possibly learned during Learning from Human Feedback, but nonetheless act as amplifiers of ongoing language change. Even if not the ultimate source of these trends, machine-generated language influences human language, which is significant in its own right. 

Either way, there is no counterfactual and we cannot definitively attribute the changes to LLM influence. Similar patterns might have emerged independently of LLMs. Therefore, additional evidence is required to support a stronger causal claim. Gaining such evidence is extremely difficult, it would require an extensive longitudinal study:\ one would need to track speakers over time, assess the degree and nature of their exposure to LLM-generated language, document their AI usage, and analyze how their lexical choices evolve. 

As others have previously observed \cite{hataya2023will,briesch2023large,alemohammad2023self}, today's language production becomes tomorrow's LLM training data, the day after tomorrow's LLM output, and eventually part of the linguistic input humans encounter. There is the possibility that a (semi-)\textit{self-contained training-loop}, as \citet{briesch2023large} put it, is forming, which comes with computationally negative consequences \cite{shumailov2023curse}.

\subsection{Limitations}

There are several limitations to our work. First, a larger dataset would be advantageous. In particular, for the baseline words, we would expect to see somewhat less variation than currently observed, suggesting that some of the variation we do observe may be due to random fluctuations in a relatively small dataset. In fact, we initially planned to work with 5 million words per time period, but we realized that this would be too limited for meaningful statistical analysis. Still, we believe that the observed increase in AI-associated words offers an accurate reflection of a genuine underlying trend. 

Second, the effects we observe likely do not reflect general population trends. Our focus on technology and science-related podcasts may amplify certain language changes – a strategic choice intended to increase the likelihood of capturing LLM-influenced language. Arguably, the results hold for a certain demographic (approximately:\ tech-affine individuals in the young to middle-aged range). However, it also means that the results are probably not representative of all spoken language. For example, an analysis of rural sermons would likely show no noticeable changes in the usage of the AI-associated words we tracked. Further, our dataset is drawn from English-language podcasts that are popular in the United States, but often feature international speakers. As such, the findings reflect general English usage in globally oriented, tech-focused discourse. Further research with more fine-grained data is needed to explore regional and varietal differences. That said, we do not think that these limitations are problematic. Our goal was to measure onset language change, which often originates within specific subgroups, before diffusing more broadly across the general speaker community. Examples of this are the rise of ``touch base'' (originating in business contexts) or ``dude'' (youth culture).

Regarding the dataset itself, some podcasts may include scripted promotional segments, which were especially difficult to isolate and remove when placed mid-episode. Additionally, podcast interviews may contain pre-scripted questions from the host; however, the ensuing discussions are, in most cases, spontaneous and unscripted. In both cases, these segments are extremely difficult to filter out automatically and would require an impractical amount of manual effort. In practice, however, they constitute only a small fraction of each episode.

Further, a qualitative analysis, linking underlying micro-level patterns to the observed macro-level trends would have nicely complemented our findings (among others, \citealp{gee2014introduction} motivates the value of qualitative analyses). Such work could be similar to \citet{krielke2024cross}, who traced the mechanisms of syntactic change through detailed case studies, and could also take recent insights from longitudinal changes in the syntactic structure of Scientific English into account (for recent insights, see e.g.\ \citealt{krielke2021relativizers,krielke2022tracing,krielke2023optimizing,chen2024syntactic}). Following up on our work, one could examine selected instances of words such as ``surpass,'' ``boast,'' and especially ``align.'' This would involve analyzing the contexts in which these terms are used, identifying recurring discourse frames \cite{fillmore2006frame}, and exploring whether particular speaker types or roles tend to produce them \cite{goffman1981forms}. Such a qualitative analysis could also help clarify developments that defied expectations:\ How were e.g.\ ``realm'' and ``groundbreaking'' used in academic writing prior to the arrival of LLMs? How has their usage developed since? And how does this contrast with their usage in spoken language? Extensive qualitative analyses constitute substantial undertakings in their own right, Extensive qualitative analyses are projects in their own right, and we must leave these for future research. 

\subsection{Broader societal implications}

Our central contribution lies in demonstrating that recent changes in language use may not be solely attributable to humans using AI models as a tool. Rather, our findings suggest changes in the human language system itself, and a plausible possibility is that AI is beginning to directly influence the human language system (an interpretation that is supported by the observed overlap between AI-overused words and those increasing in human usage). This has broader relevance for ongoing debates around model alignment. There is a gap between model language behavior and human language behavior, which raises important questions about how to interpret this gap and about potential societal consequences of such misalignments. While we are concerned with (mis)alignment of language behavior, this, of course, is directly linked to alignment discussions on social or ethical values, with the issues and mechanisms of misalignment manifestation and propagation being very similar. 

One critical aspect we identify is what we call the ``seep-in'' effect. There is a gap between AI behavior and human values and behavior, and based on our results, when it comes to language, there is a plausible possibility that AI linguistic patterns are beginning to \textit{seep into} the human language system. That is, a word or expression that may not have reflected a speaker's own linguistic preference may nonetheless become part of their language system, simply through repeated exposure \cite{tomasello2005constructing,bybee2006usage} -- exposure now increasingly impacted by AI-generated language (as well as by changes in the writing process itself, which might become increasingly AI-assisted \cite{floridi2025distant}). While such a seep-in may seem relatively benign in the case of lexical choices, it raises more serious concerns when considered in relation to political or social beliefs of misaligned models or even models by malicious actors. 

Work on the features of AI-associated language also has potential relevance for AI detection (for ongoing efforts, see e.g.\ \citealt{jin2025trapdoc,koziev2025detecting,schmalz-tack-2025-gptzeros}), a task that remains challenging \cite{sadasivan2023can, weber2023testing}. Human language is remarkably complex (recently, e.g.\ \citealt{wang2023linguistic,wu2024perplexing,zhang2024cross,clark2025relationship}), which could be a contributing factor to this challenge. Advancing detection methods will require both high-quality tools and robust benchmark datasets (recent advances include \citealt{warstadt2020blimp,srivastava2023beyond,lucking2024dependencies,zhang2024cross,weissweiler2025linguistic,jumelet2025multiblimp}).

One might ask whether these linguistic developments are fundamentally new. As noted in the Introduction, technology and language have long had a reciprocal relationship. From the printing press to the rise of digital media, linguistic behavior has evolved alongside changes in communication technology. What distinguishes the current moment is the pace and scale of language change. If the adoption of Large Language Models continues to grow globally, then a broader and potentially homogenizing effect on language is plausible (also cf.\ \citealp{bender2021dangers}). Over the coming decades, we may see convergence around an AI-influenced register, with all its linguistic implications. 

An alternative interpretation is that these lexical shifts reflect a trajectory of natural language change, with AI serving as an accelerator or amplifier. While this cannot be ruled out, it is difficult to test empirically. Anecdotally, we observe that human choice still plays a role. For instance, the word ``delve,'' which saw widespread usage in academic writing in 2023 and 2024 (as per Introduction), is falling out of favor. In Q1 2025, occurrences of ``delve'' in PubMed abstracts dropped to about 15 occurrences per one million words, from about 22 occurrences per one million in the previous year. Thus, there is a \textit{delve-pushback}, a reversal in language usage, over a very short period of time, for which we can only speculate about the underlying motivation. It may be that speakers became aware of how ``delve'' had seeped into their language system and reacted against it; and/or ``delve'' may have become associated with AI-generated language and, in some contexts, particularly scientific writing, now carries a degree of stigma. 

What is certainly new, however, is the problem of human-authorship indeterminacy. For an increasing number of texts, it is no longer possible to determine whether the content was genuinely produced by a human. This has profound implications for linguistic research and computational linguistics in particular, where language data has traditionally served as a proxy for human cognition and behavior. Our study exemplifies this disruption:\ what was once a relatively straightforward task (measuring trends in human language production), now required substantial methodological effort to ensure the validity of the data. As authorship indeterminacy becomes more widespread, it will complicate not only linguistic research, but also any scientific field that relies on natural language as a proxy for human behavior and cognition.

\section{Conclusion}

Our work contributes to research on AI's impact on human language, specifically addressing whether recent changes in language use stem from direct AI tool usage or whether genuinely human-produced language is beginning to follow patterns associated with AI language, particularly the overuse of certain words such as ``delve'' and ``intricate.'' In this view, such words may be gaining prominence in people's mental language systems, leading to increased usage over time. 

To investigate this, we constructed a dataset of unscripted spoken language as a proxy for spontaneous spoken language. We divided the dataset into pre- and post-2022 periods and examined changes in the frequency of words identified in the literature as commonly overused by Large Language Models. Our findings reveal a mostly positive trend in the usage of these AI-associated words, in contrast to a balanced pattern among baseline words, where increases and decreases are roughly evenly split. While these early findings are promising, larger datasets and further analysis are needed to substantiate the observed trends. 

A link between changes in human language usage and AI is plausible, given the observed overlap in affected vocabulary. However, we leave open the question of whether the observed increases reflect AI-induced language change or natural language change. This distinction is difficult to determine and remains an ongoing challenge. However, it is a crucial one. If the patterns align with established processes of language change, viz.\ human preferences driving human language, then there may be little cause for concern. But if the language favored by AI reflects preferences not rooted in natural human usage, then we are witnessing a moment of model misalignment. In particular, the overuse of certain words by LLMs is widely believed to stem from the training procedure known as Learning from Human Feedback. While Learning from Human Feedback is designed to align models with human preferences, it may in practice amplify stylistic features that do not reflect the expectations of most users. In this sense, upstream alignment choices can introduce downstream linguistic biases. This matters not only for understanding lexical shifts but also because the mechanisms of linguistic (mis)alignment closely mirror broader concerns of (mis)alignment in AI ethics and fairness.

Of particular concern is what one might call the ``seep-in'' effect, the subconscious adoption of AI-driven linguistic or even ethical tendencies, even when these do not reflect the speaker's original preferences. If such preferences gradually permeate human communication, this could mark a novel and potentially profound shift in the trajectory of language evolution. Lastly, as our study exemplifies, AI is not only reshaping language use, it is disrupting how we study language itself. For many written texts, human authorship is no longer easily determined, which makes it increasingly difficult to infer human behavior from language patterns. 

\appendix

\section{Appendix A:\ Computing resources}
\begin{center}
  \rule{7cm}{0.5pt}
\end{center}
\label{sec:appendix-a}

This appendix details the computing environment used to run the scripts and analyses in this study.

\subsection*{System Specifications}
\begin{itemize}
  \item \textbf{Machine:} 2024 Thelio Custom Machine
  \item \textbf{Operating System:} Ubuntu 24.04.1 LTS
  \item \textbf{CPU:} Intel Core i7-14700K
  \item \textbf{GPU:} NVIDIA GeForce RTX 3090
  \item \textbf{RAM:} 128 GB
\end{itemize}

\begin{center}
  \rule{7cm}{0.5pt}
\end{center}

\subsection*{Software Environment}
\begin{itemize}
  \item \textbf{Python Version:} \texttt{3.12.3}
  \item \textbf{Key Libraries:}
  \begin{itemize}
    \item \texttt{openai-whisper == 20240930}
    \item \texttt{spacy == 3.8.2}
    \item \texttt{scipy == 1.15.2}
  \end{itemize}
\end{itemize}

\begin{center}
  \rule{7cm}{0.5pt}
\end{center}

\subsection*{Runtime Notes}
\begin{itemize}
  \item \textbf{Total transcription time / Whisper runtime:} 92 hours
  \item \textbf{spaCy tagging time:} 2 hours
  \item \textbf{Average analysis runtime:} 4 minutes per 1 hour of transcription
\end{itemize}

\subsection*{Additional Notes}
This environment uses a 2024 Thelio Custom Machine equipped with an Intel Core i7-14700K processor and an NVIDIA GeForce RTX 3090 GPU.

\begin{center}
  \rule{7cm}{0.5pt}
\end{center}

\section{Appendix B:\ Full results for the 117 baseline words}
\label{sec:appendix-b}

Out of the 117 synonyms from the literature, 87 did occur in our dataset; the results for these can be found in Tables~\ref{tab:synos1} and \ref{tab:synos2} below. The 30 words not occurring in our dataset are:\ systematize\_VERB, harmonize\_VERB, underscore\_VERB, laudable\_ADJ, praiseworthy\_ADJ, noteworthy\_ADJ, meritorious\_ADJ, understanding\_ADJ, grasping\_ADJ, perceiving\_ADJ, interpreting\_ADJ, discerning\_ADJ, accentuate\_VERB, highlighting\_ADJ, underscoring\_ADJ, stressing\_ADJ, prioritizing\_ADJ, collected\_ADJ, acquired\_ADJ, obtained\_ADJ, assembled\_ADJ, amassed\_ADJ, intricateness\_NOUN, inestimable\_ADJ, priceless\_ADJ, immeasurable\_ADJ, diligent\_ADJ, scrupulous\_ADJ, tactically\_ADV, methodically\_ADV.

\begin{table*}[t]
\centering
\begin{tabular}{|l|c|c|c|c|}
\hline
\textbf{Word} & \textbf{Change (\%)} & \textbf{OPM Pre-2022} & \textbf{OPM Post-2022} & \textbf{p \ensuremath{\leq} 0.05} \\
\hline
irreplaceable\_ADJ & 275.86 & 0.27 & 1.03 & False \\
eminent\_ADJ       & 114.78 & 0.64 & 1.38 & False \\
territory\_NOUN    & 93.88  & 14.47 & 28.05 & \textbf{True} \\
exhibit\_NOUN      & 93.46  & 1.56 & 3.01 & \textbf{True} \\
admirable\_ADJ     & 87.93  & 1.19 & 2.24 & False \\
capability\_NOUN   & 81.71  & 30.40 & 55.24 & \textbf{True} \\
multifaceted\_ADJ  & 67.05  & 0.82 & 1.38 & False \\
probable\_ADJ      & 57.86  & 2.29 & 3.61 & False \\
notable\_ADJ       & 53.13  & 2.47 & 3.79 & False \\
achievable\_ADJ    & 50.35  & 0.92 & 1.38 & False \\
assemble\_VERB     & 46.48  & 6.23 & 9.12 & \textbf{True} \\
thorough\_ADJ      & 43.19  & 1.92 & 2.75 & False \\
reinforce\_VERB    & 38.37  & 8.33 & 11.53 & \textbf{True} \\
novel\_ADJ         & 38.22  & 15.75 & 21.77 & \textbf{True} \\
grasp\_VERB        & 27.90  & 6.59 & 8.43 & False \\
complexity\_NOUN   & 27.71  & 55.58 & 70.98 & \textbf{True} \\
visionary\_ADJ     & 21.60  & 1.56 & 1.89 & False \\
enhancement\_NOUN  & 20.81  & 1.92 & 2.32 & False \\
prospect\_NOUN     & 17.14  & 6.68 & 7.83 & False \\
stress\_VERB       & 16.98  & 8.97 & 10.50 & False \\
integrate\_VERB    & 16.90  & 27.01 & 31.58 & \textbf{True} \\
sophisticated\_ADJ & 15.68  & 22.98 & 26.59 & False \\
coordinate\_VERB   & 15.32  & 8.06 & 9.29 & False \\
elaborate\_ADJ     & 15.18  & 5.68 & 6.54 & False \\
deliberately\_ADV  & 14.85  & 8.24 & 9.46 & False \\
remarkable\_ADJ    & 14.35  & 24.91 & 28.48 & False \\
critical\_ADJ      & 11.46  & 51.64 & 57.56 & False \\
vital\_ADJ         & 10.55  & 6.23 & 6.88 & False \\
original\_ADJ      & 9.60   & 58.88 & 64.53 & False \\
highlight\_VERB    & 9.54   & 16.57 & 18.15 & False \\
improvement\_NOUN  & 9.20   & 23.72 & 25.90 & False \\
complex\_ADJ       & 7.57   & 86.62 & 93.18 & False \\
vigilant\_ADJ      & 6.49   & 1.37 & 1.46 & False \\
showcase\_VERB     & 6.49   & 1.37 & 1.46 & False \\
convoluted\_ADJ    & 5.71   & 1.47 & 1.55 & False \\
possibility\_NOUN  & 4.69   & 77.83 & 81.48 & False \\
innovative\_ADJ    & 4.64   & 8.06 & 8.43 & False \\
merit\_NOUN        & 4.09   & 5.95 & 6.19 & False \\
complication\_NOUN & 2.78   & 5.86 & 6.02 & False \\
renowned\_ADJ      & 1.80   & 1.10 & 1.12 & False \\
pivotal\_ADJ       & 0.45   & 2.66 & 2.67 & False \\
inventive\_ADJ     & 0.23   & 1.37 & 1.38 & False \\
\hline
\end{tabular}
\caption{Increases in word frequency (occurrences per million; OPM) for the baselines synonyms, for pre-2022 vs.\ post-2024, including p-values (part 1).}
\label{tab:synos1}
\end{table*}

\begin{table*}[t]
\centering
\begin{tabular}{|l|c|c|c|c|}
\hline
\textbf{Word} & \textbf{Change (\%)} & \textbf{OPM Pre-2022} & \textbf{OPM Post-2022} & \textbf{p \ensuremath{\leq} 0.05} \\
\hline
development\_NOUN  & -0.81  & 70.87 & 70.29 & False \\
possible\_ADJ      & -1.05  & 260.50 & 257.77 & False \\
research\_VERB     & -1.26  & 16.21 & 16.00 & False \\
distinction\_NOUN  & -1.94  & 33.60 & 32.95 & False \\
understand\_VERB   & -2.36  & 587.84 & 573.97 & False \\
presentation\_NOUN & -3.10  & 11.72 & 11.36 & False \\
explore\_VERB      & -3.31  & 69.41 & 67.11 & False \\
demonstration\_NOUN & -4.19 & 9.34 & 8.95 & False \\
precise\_ADJ       & -4.97  & 16.21 & 15.40 & False \\
appreciate\_VERB   & -5.42  & 83.69 & 79.16 & False \\
discipline\_NOUN   & -6.79  & 22.80 & 21.25 & False \\
exhibit\_VERB      & -7.50  & 5.86 & 5.42 & False \\
probe\_VERB        & -8.72  & 6.41 & 5.85 & False \\
theoretical\_ADJ   & -8.98  & 38.00 & 34.59 & False \\
sphere\_NOUN       & -9.13  & 16.66 & 15.14 & False \\
display\_VERB      & -9.95  & 6.59 & 5.94 & False \\
perceive\_VERB     & -10.06 & 29.94 & 26.93 & False \\
consciously\_ADV   & -11.07 & 5.13 & 4.56 & False \\
key\_ADJ           & -12.89 & 60.25 & 52.48 & \textbf{True} \\
highlight\_NOUN    & -14.20 & 4.21 & 3.61 & False \\
progress\_NOUN     & -14.74 & 67.21 & 57.30 & \textbf{True} \\
compile\_VERB      & -14.88 & 7.78 & 6.63 & False \\
field\_NOUN        & -16.01 & 232.02 & 194.88 & \textbf{True} \\
valuable\_ADJ      & -16.47 & 37.08 & 30.97 & \textbf{True} \\
feature\_NOUN      & -16.91 & 112.35 & 93.35 & \textbf{True} \\
forum\_NOUN        & -17.57 & 5.22 & 4.30 & False \\
essential\_ADJ     & -19.12 & 28.93 & 23.40 & \textbf{True} \\
aggregate\_VERB    & -19.15 & 3.94 & 3.18 & False \\
likelihood\_NOUN   & -19.72 & 9.43 & 7.57 & False \\
display\_NOUN      & -20.18 & 8.52 & 6.80 & False \\
purposefully\_ADV  & -22.62 & 1.56 & 1.20 & False \\
crucial\_ADJ       & -24.56 & 19.05 & 14.37 & \textbf{True} \\
investigate\_VERB  & -25.01 & 27.65 & 20.74 & \textbf{True} \\
synchronize\_VERB  & -26.46 & 2.11 & 1.55 & False \\
excellence\_NOUN   & -29.53 & 5.13 & 3.61 & False \\
collect\_VERB      & -30.71 & 59.97 & 41.56 & \textbf{True} \\
amass\_VERB        & -30.76 & 1.74 & 1.20 & False \\
distinguished\_ADJ & -31.96 & 2.66 & 1.81 & False \\
examine\_VERB      & -32.00 & 13.92 & 9.46 & \textbf{True} \\
feasible\_ADJ      & -34.02 & 8.61 & 5.68 & \textbf{True} \\
probability\_NOUN  & -34.75 & 52.74 & 34.42 & \textbf{True} \\
outstanding\_ADJ   & -36.02 & 4.30 & 2.75 & False \\
domain\_NOUN       & -52.32 & 55.40 & 26.41 & \textbf{True} \\
sophistication\_NOUN & -53.98 & 4.49 & 2.06 & \textbf{True} \\
convolution\_NOUN  & -79.37 & 7.51 & 1.55 & \textbf{True} \\
\hline
\end{tabular}
\caption{Decrease in word frequency (occurrences per million; OPM) for the baselines synonyms, for pre-2022 vs.\ post-2024, including p-values (part 2).}
\label{tab:synos2}
\end{table*}

\bibliography{aaai25}

\end{document}